\documentclass[[article,11pt,onecolumn,final]{IEEEtran}
\IEEEoverridecommandlockouts
\usepackage{url}
\usepackage{subfig}
\usepackage{cite}
\usepackage{amsmath,amssymb,amsfonts}
\usepackage{algorithmic}
\usepackage{graphicx}
\usepackage{textcomp}
\usepackage{adjustbox}
\usepackage{graphicx}
\usepackage{xcolor}
\usepackage[textsize=tiny,textwidth=1.4cm]{todonotes}
\usepackage{mathtools,lipsum, nccmath,cuted}
\usepackage{siunitx}
\usepackage{textcomp}

\usepackage{comment}
\usepackage{makecell}
\usepackage{algorithm}
\usepackage{algorithmic}
\usepackage{gensymb}
\usepackage{ulem}
\usepackage{MnSymbol}
\usepackage{colortbl}
\newtheorem{definition}{Definition}
\usepackage{enumitem}
\definecolor{Gray}{gray}{0.925}
\usepackage[textsize=tiny]{todonotes}

\graphicspath{ {figures/} }
\def\BibTeX{{\rm B\kern-.05em{\sc i\kern-.025em b}\kern-.08em
    T\kern-.1667em\lower.7ex\hbox{E}\kern-.125emX}}
    
\usepackage{pgfplots}
\usepgfplotslibrary{colorbrewer}
\pgfplotsset{compat = 1.14, cycle list/Set1-8} 
\usetikzlibrary{pgfplots.statistics, pgfplots.colorbrewer,pgfplots.dateplot,
        shadows,
        matrix} 

\usepackage{pgfplotstable}
\usepackage{filecontents}
\usepackage{graphicx}
\pgfplotsset{compat=1.14}

\definecolor{blueLine}{RGB}{57,106,177}
\definecolor{blueFill}{RGB}{114,147,203}
\definecolor{redLine}{RGB}{204,37,41}
\definecolor{greenline}{RGB}{0,250,0}
\definecolor{blackLine}{RGB}{0,0,0}
\definecolor{goldLine}{RGB}{160,82,45}

\usepackage[font=footnotesize]{caption}
\begin{document}

\title{Systematic design space exploration by learning the explored space using Machine Learning }

\author{\IEEEauthorblockN{Avinash Kumar, Anish Kumar, Sumit Sharma, Surjeet Singh, Kumar Vardhan*}
\IEEEauthorblockA{
\textit{\\ *University of Cincinnati, USA } 
}
}

\maketitle

\section{Abstract}
\textit{Current practice in parameter space exploration in euclidean space is dominated by randomized sampling or design of experiment methods. The biggest issue with these methods is not keeping track of what part of parameter space has been explored and what has not. In this context, we utilize the geometric learning of explored data space using modern machine learning methods to keep track of already explored regions and samples from the regions that are unexplored. For this purpose, we use a modified version of a robust random-cut forest along with other heuristic-based approaches. We demonstrate our method and its progression in two-dimensional Euclidean space but it can be extended to any dimension since the underlying method is generic.}

\section{Introduction}
Design space exploration\cite{box1959design} has a major role in both design optimization\cite{martins2013multidisciplinary} as well as surrogate modeling\cite{bouhlel2019python}. It is the process of discovering and evaluating the manifold of the function or process under consideration. The goal in the case of design optimization is to find the parameter\cite{adeli1994advances} that performs the best on the performance metrics and meets all requirements. On the other hand, in the case of surrogate modeling the goal is to learn a data-driven alternative model that is cheaper to evaluate and acts as a proxy of the function under consideration. In both cases, a systematic exploration of design space is required especially in high-dimensional design space. 
Current practice in design space exploration is done by either random sampling or traditional design of experiment methods\cite{anderson2018design}. Random sampling generally relies on pseudo-random compute-generated sequences\cite{sobol1999pseudo}. In DoE\cite{fisher1936design} methods, factorial-based\cite{hicks1964fundamental} sampling is the most used. In Latin hypercube-based sampling (LHC)\cite{mckay1992latin} sampling and its flavor is most used\cite{morris1995exploratory}. Extension of these factorial methods  relies on embedding factorial or fractional factorial design with points that are augmented with a group of specific shape points with some invariant properties (like orthogonality, rotatibility, etc) that allow easy estimation of the response surface\cite{box1992experimental,box2007response}.
\\
All these methods do not keep track of what part of the design space is explored and what part is not explored. Consequently, there is a high likelihood of sample collision. Sample collision is a waste of resources and it becomes more problematic in high-dimensional design search space. To address this, in this work we developed a method that relies on geometric summarization of explored space in sketches and use the $\epsilon$-hyper-ball based heuristic to select new samples for exploration. It is an iterative method and at each iteration, we select a batch of samples and learn the explored space using robust random cut forest\cite{guha2016robust} and then we find the peripheral design points. These peripheral design points are used to further create an $\epsilon$-hyper-ball and then samples from it. We showed multiple experiments and shows how $\epsilon$ can be used to control the step sizing or density of sampling. 
\\
Accordingly, in this work, we developed a machine learning-based method for systematic sampling of design space.

\section{The problem}
In this section, we formalize our problem. For this purpose, we first formalize the euclidean space learning problem.  On a given data set $D \in R^m$ where $D=\{d_1,d_2, ..., D_n\}$ is the already sampled data points, the design space learning problem is finding an operator $\Omega$, where  $\Omega$ maps the dataset from Euclidean space to a structured trainable parametric or non-parametric model ($M$) Once trained, we want our prediction model to work in a subspace of input space and detect the input data which is not part of training data subspace  by raising a flag for taking some corrective measure. 
$$ \Omega: D \mapsto M $$
Once trained, we want to find the points that are on the periphery of the design space in Euclidean space using the trained model ($M$). The model $M$ should not only be able to predict what points are in the learned sub-space and what are not but also be able to give an inference about the points on the periphery. Let $P$ be the peripheral points. Then we extend these peripheral points by sampling beyond these peripheral points and add these points $D_{new}$ to already explored data points $D$.

\section{Approach}
In our approach for learning the explored space ($T$) occupied by $D$ in to the trained model $M$, we  rely on learning the cluster of design sub-space ($T$) 's shape in Euclidean space to a data-structure ($S$). The motivation behind learning the cluster's shape into a data structure is to abstract the information from metric space in a structured manner such that computer and related algorithms can be efficiently deployed for inference.  If $D =\{d_1, d_2, ... , d_n\}$ are set of datapoints such that $d_i \in R^m $. For the purpose of Out-of-Distribution detection, following requirements are imposed on this data-structure($S$):  
\begin{enumerate}
    \item $S$ should represent the cluster of data in a structured way. 
    \item Relationships($\psi$) between the data points in metric space must be preserved in this data structure. i.e 
    $$\psi\{T(d_k,d_l)\}\approx \psi\{S(d_k,d_l)\} $$
    \item Relationship($\phi$) of a data with the explored space should be encoded in simple quantitative measure. i.e. $\phi(T,d_k)$ can be measured as a scalar value in the data-structure($S$). 
    \item Any modification and inference using this data structure (S) should be computationally cheap.
    \item All the above properties should be compatible with streaming data.
 \end{enumerate}
To represent our cluster in an organized manner, we choose the RRCF\cite{guha2016robust} as our choice of data structure.   Robust Random Cut Forest (RRCF) can be defined as follow :
\begin{definition}
  Robust Random Cut Tree on set of data point $D=\{d_1,d_2,...,d_n\}$ can be generated by following procedure: 
 \end{definition}
 \begin{enumerate}
     \item $r_i= max_{X \in D}(X_i)-min_{X \in D}(X_i) \;\forall i\in m$
     \item $p_i= \dfrac{r_i}{\sum_{i=1}^{i=m}r_i} \;\forall i$
     \item select a random dimension $i$ with probability proportional to $p_i$
     \item $choose \; x_i \mid x_i \sim Uniform(max(X_i)-min(X_i))$
     \item $ D_1=\{X \mid X\in D ,X_i \leq x_i \}$
     \item $D_2= D \setminus D_1$
 \end{enumerate}
 $recurse\; on\; D_1\; and\; D_2\; until\; D_i \geq 1 $ 

Robust Random cut Forest is an ensemble of various RRCT. 
We need a distance-preserving embedding in the data structure since the chosen relationship ($psi$) between data points in metric space is represented by the $L p$ distance between data points.  For this reason, the weight of the least common ancestor of two datapoints $d k$ and $d l$ is used to establish the tree distance between these points in the data structure ($S$)\cite{guha2016robust}. Following Johnson-Landatrauss lemma\cite{lindenstrauss1984extensions} the tree distance can be bounded from at-least  $L_1 (d_k,d_l)$ to maximum $ O(d* log|k|/L_1(d_k,d_l))$.
Consequently, a point will remain at least as far from other points in a random cut tree if it is far from them in metric space.  
Displacement, which is an estimation of the change in model complexity (summation of leave's depth) before and after adding a given point $x$ in tree data structure, can be used to convey the relationship ($phi$) of a data point with the cluster in a straightforward quantitative manner.

\begin{figure}[h!]
        \centering
        \includegraphics[width=0.7\textwidth]{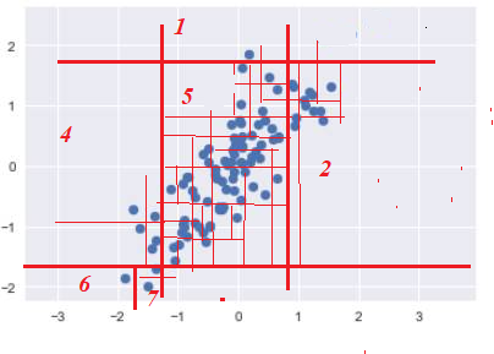}
        \caption{A robust random cut tree on sample data}
        \label{fig:cut tree}
    \end{figure}
\begin{figure}[h!]
        \centering
        \includegraphics[width=0.6\textwidth]{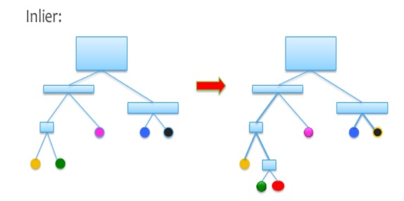}
        \caption{An inlier data point and its position in the tree. }
        \label{fig:inlier}
    \end{figure}

By using these RRCF\cite{vardhan2022reduced}, we are interested in learning a subspace in parameter space and also finding the data points that are on the periphery. At the start, we select a batch of samples randomly (call $D$), called the warm-up stage. These random samples are used to train various  RRCT. The cut space created by an RRCT is shown in figure \ref{fig:cut tree}. The ensemble of RRCT that forms an RRCF is trained on this initial data set. This RRCF represents our model $M$.  Our next goal is to find the points on the periphery. For this purpose, we use the metric \textbf{displacement} that is defined above. The insertion of a point in from cluster called inlier has a high likelihood that once inserted into the RRCF, it will get inserted at  the bottom part of the tree (refer to figure \ref{fig:inlier}) and consequently the change in displacement would be lesser than the point that is an outlier. 

\begin{figure}[h!]
        \centering
        \includegraphics[width=0.6\textwidth]{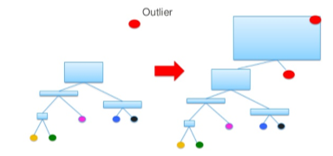}
        \caption{An outlier data point and its position in tree.}
        \label{fig:outlier}
\end{figure}

An outlier has a high likelihood that it gets attached at the initial branches of the tree and consequently increases the displacement to a maximum amount (refer to figure \ref{fig:outlier}). We capitalize on it and run this displacement calculation on all points in the data set and select the points that have maximum displacement. For a batch of samples, we select the batch of point order in descending order based on displacement. Collection of These selected points is the peripheral point set $P$. 
The next step is to explore and add new points. For this purpose, we create an $\epsilon$-hyperball with a center on these $P$ points. The $\epsilon$ is a hyper-parameter and controls the sparsity and density of the design points. Bigger $\epsilon$ represents the sparser selection of sample and vice-versa.  We select one sample from each hyper-ball, evaluate it and label it as newly sampled data set $D_{new}$. The data set $D$ is modified as 
 $$D \leftmapsto D \cup D_{new}$$
we again train our RRCF on the data set $D$ and repeat the process again. 

\section{Results}
   \begin{figure}[h!]
        \centering
        \includegraphics[width=0.9\textwidth]{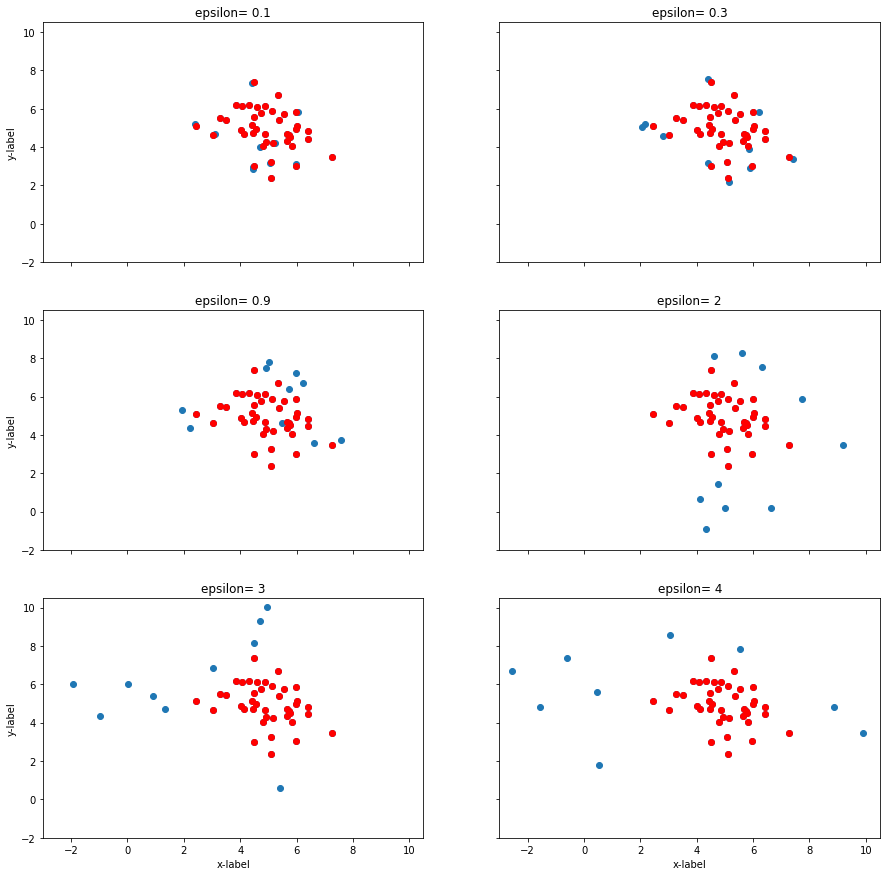}
        \caption{Result of different epsilon using our algorithm. Blue points represent the newly selected points and red points are points that are already explored.}
        \label{fig:diff_epsilon}
    \end{figure}
\begin{figure}[h!]
        \centering
        \includegraphics[width=0.85\textwidth]{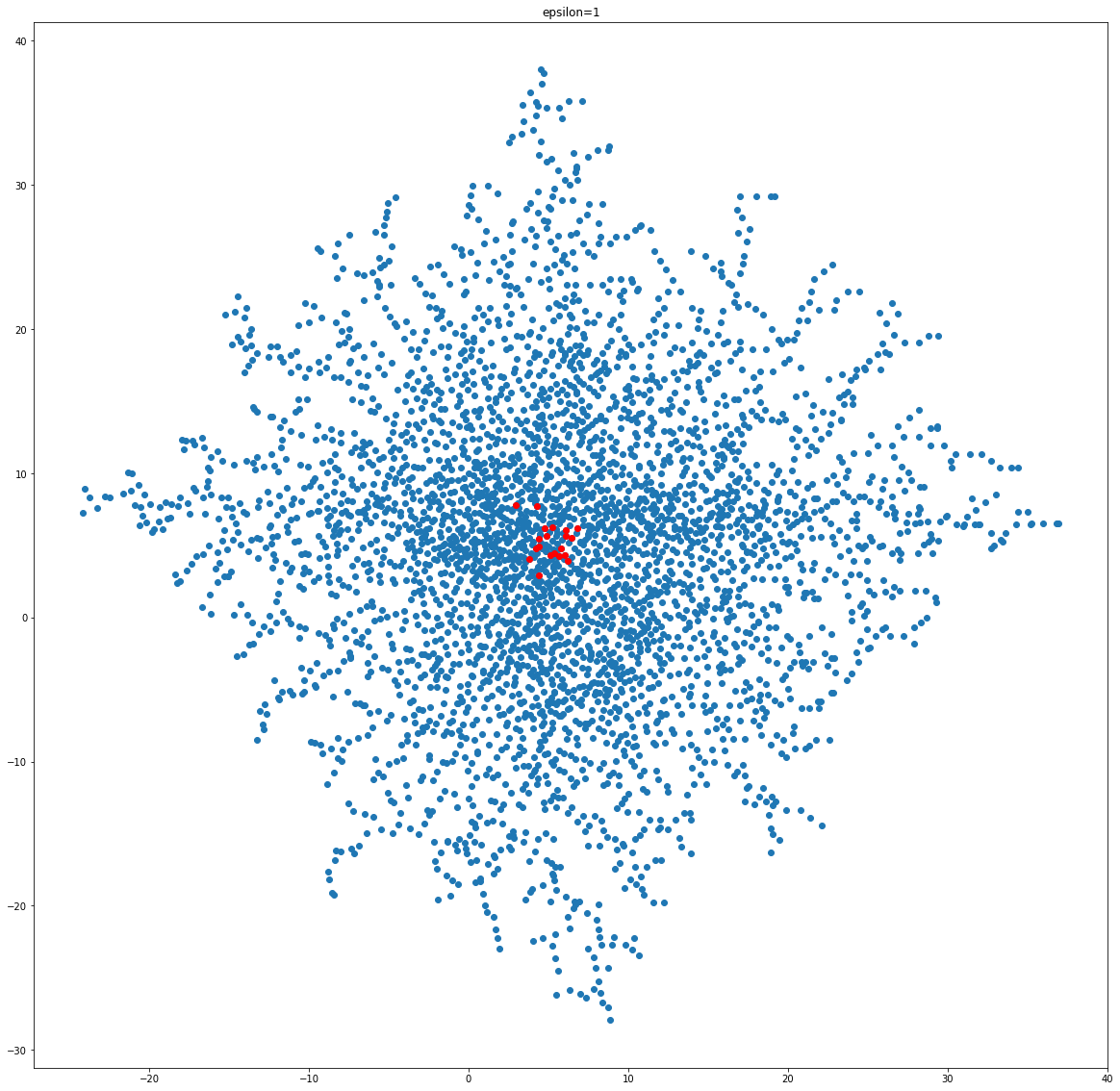}
        \caption{Result on a large number of samples on a fixed epsilon using our algorithm.}
        \label{fig:bigone}
    \end{figure}

To demonstrate and visual explanation, we test our approach for parameter space exploration in 2D space. Figure \ref{fig:diff_epsilon} shows the effect of $\epsilon$ on the sparsity of the design space exploration process at one iteration. By controlling the $\epsilon$, we control the hyper-ball size created around the peripheral points $P$. At $\epsilon=0.1$, the selected samples are closest to the already explored region, and by increasing the values of $\epsilon$, the selected samples become more and more separated from the already explored data cluster. With $\epsilon=4$, the selected samples are widely separated from the already sampled cluster. 
\\
In the next experiment, we let our algorithm run for 2000 iterations with $50$ samples per iteration. The value of $\epsilon$ is kept fixed and we are only interested in observing does the algorithm work as we expected when we let it run for a longer iteration. The red samples in figure \ref{fig:bigone} are the initial samples chosen randomly and the blue points are the points selected by the  algorithm. It can be observed that algorithm performs as expected.

\section{Related work}
Three key elements in engineering include model-based design (\cite{neema2019design}), model-based control (\cite{brosilow2002techniques}), and model-based optimization (\cite{vardhan2019modeling}). Machine learning and AI is affecting all three fields like for control\cite{duan2016benchmarking,neema2019web,vardhan2023search,wu2019machine,duriez2017machine}, for modeling\cite{audet2000surrogate}, for prediction of complex systems\cite{vardhan2021machine,volk2020biosystems,vardhan2022deepal,mazurenko2019machine,vardhan2022data,kumar2023malaria, vardhan2022deep} and for optimization\cite{vardhan2023search,vardhan2023constrained,frazier2018tutorial,vardhan2023fusion}. Parameter space exploration is crucial for all these applications. Systematic exploration has benefits as it can avoid sample collision and would be the most useful in unstructured regions of design space. The early work in design space exploration is done by Fisher \cite{fisher1960design}. Later Box and Wilson gave Box-Wilson Central Composite Design of an experiment. Later Joan Fisher's box revisited the design of the experiment work. Till now, most of the DoE methods rely on factorial methods. Latin hypercube \cite{loh1996latin} is one the most used method. First introduced by \cite{mckay1992latin}, it relies on the factorial separation of space, with the goal of non-colliding sample selection. LHC relies on creating a square grid containing sample positions in a Latin square if (and only if) there is only one sample in each row and each column. Morris and Mitchell \cite{morris1995exploratory} added a space-filling criterion to the vanilla LHC with the optimization objective of maximizing the shortest distance between the points. 
Eriksson et al \cite{eriksson2000design,kang2010approach} provide a comprehensive tutorial on all these methods. 
On the Machine learning front, the AL methods are being used for complex control design \cite{ben2016testing, vardhan2021rare}, computer vision \cite{krizhevsky2017imagenet,lecun2015lenet,poostchi2018image}.

\section{Conclusions}
In this work, we showed a state-of-the-art approach to systematic exploration. By learning the already explored space and the periphery data points it is possible to sample new points in euclidean space. By doing this we can systematically explore the design space. In all these cases, by controlling a hyper-parameter $\epsilon$, we can control the step or spread of the sampling and exploration process. The future work would be the inclusion of constraint handling during exploration and application in real-world problems.
\bibliographystyle{IEEEtran}
\bibliography{bibliography}

\end{document}